\documentclass[10pt,twocolumn,letterpaper]{article}

\usepackage{iccv}
\usepackage{times}
\usepackage{epsfig}
\usepackage{graphicx}
\usepackage{amsmath}
\usepackage{amssymb}
\usepackage{amsthm,amsmath,epsfig,mathtools}
\usepackage{algorithm,algpseudocode}
\usepackage{multirow}
\usepackage{tabls}
\usepackage{wrapfig}

\graphicspath{{Figures/}}

\renewcommand{\vec}[1]{\mathbf{#1}}

\usepackage[pagebackref=true,breaklinks=true,letterpaper=true,colorlinks,bookmarks=false]{hyperref}

\iccvfinalcopy 

\def\iccvPaperID{7} 

\ificcvfinal\fi
\begin{document}

\title{Sequential Score Adaptation with Extreme Value Theory \\ for Robust Railway Track Inspection}

\author{Xavier Gibert\\
University of Maryland\\
College Park, MD\\
{\tt\small gibert@umiacs.umd.edu}
\and
Vishal M. Patel\\
Rutgers University\\
Piscataway, NJ\\
{\tt\small vishal.m.patel@rutgers.edu}
\and
Rama Chellappa\\
University of Maryland\\
College Park, MD\\
{\tt\small rama@umiacs.umd.edu}
}

\maketitle

\begin{abstract}
Periodic inspections are necessary to keep railroad tracks in state of good repair and prevent train accidents. Automatic track inspection using machine vision technology has become a very effective inspection tool. Because of its non-contact nature, this technology can be deployed on virtually any railway vehicle to continuously survey the tracks and send exception reports to track maintenance personnel. However, as appearance and imaging conditions vary, false alarm rates can dramatically change, making it difficult to select a good operating point. In this paper, we use extreme value theory (EVT) within a Bayesian framework to optimally adjust the sensitivity of anomaly detectors. We show that by approximating the lower tail of the probability density function (PDF) of the scores with an Exponential distribution (a special case of the Generalized Pareto distribution), and using the Gamma conjugate prior learned from the training data, it is possible to reduce the variability in false alarm rate and improve the overall performance. This method has shown an increase in the defect detection rate of rail fasteners in the presence of clutter (at PFA 0.1\%) from 95.40\% to 99.26\% on the 85-mile Northeast Corridor (NEC) 2012-2013 concrete tie dataset.
\end{abstract}

\section{Introduction}

In sequential inspection problems, such as visual railway track inspection, a video feed is streamed from one or more cameras to a detection system, and we are interested in designing a detector that can find abnormal patterns in such data. There is a limit to the number of false alarms that the operator can handle, so it is necessary to select the optimal operating point at which the false alarm rate does not exceed such limit. Indeed, most of the data that an autonomous inspection vehicle will collect will be discarded without anyone ever looking at it. Therefore, an excessively high false alarm rate will result in a waste of storage space and bandwidth. The only relevant images are the ones that correspond to unexpected patterns, so we are actually interested in finding such anomalous patterns.

\begin{figure}[t]
\begin{center}
  \includegraphics[trim=0mm 4mm 0mm 9mm, clip=true, width=3.25in]{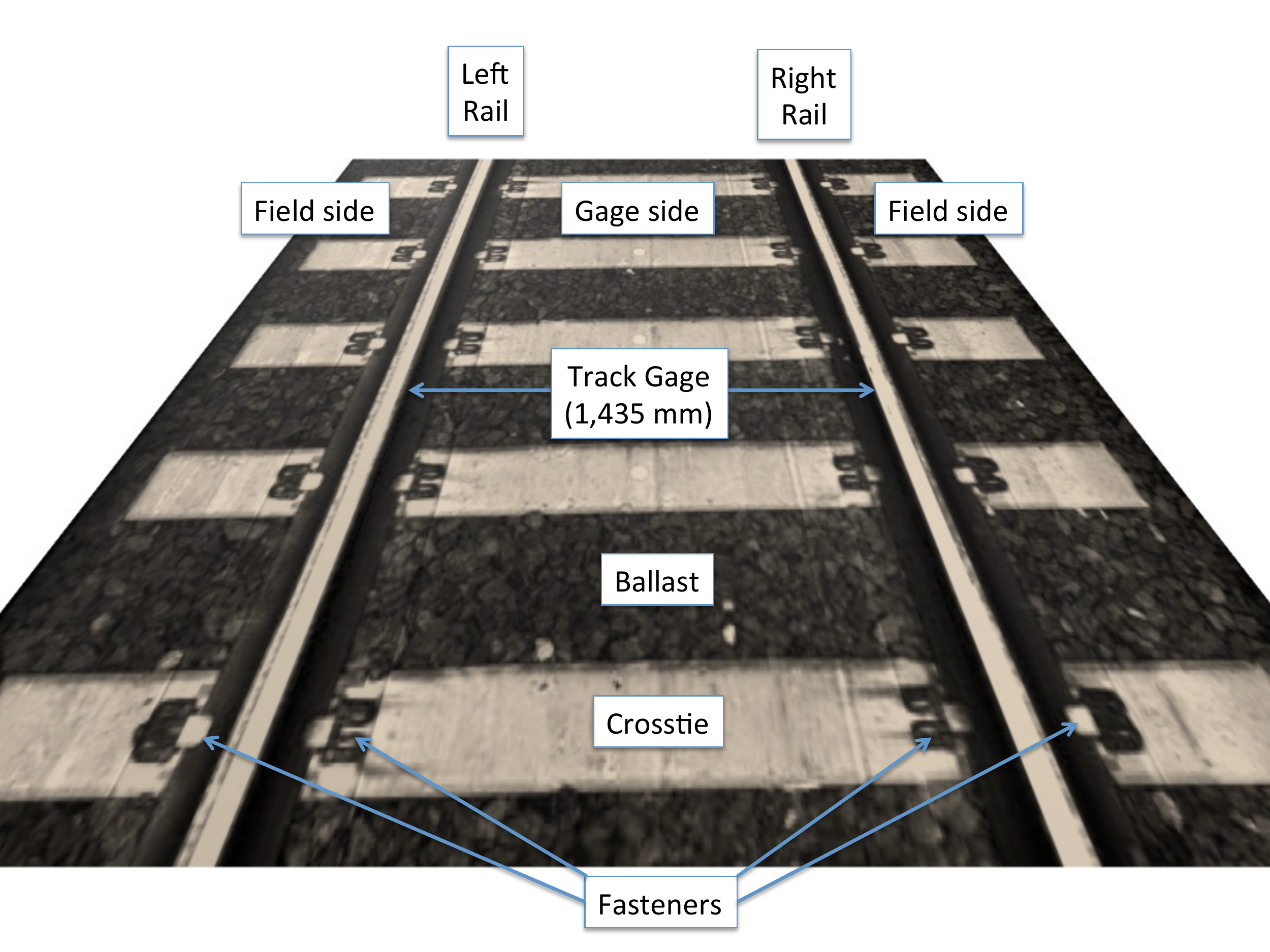}
\end{center}\vskip-10pt
\caption{ \label{fig:definitions}
Definition of basic track elements.}
\end{figure}

\begin{figure}[tp]
\begin{center}
\begin{tabular}{c c c c c}
  2.8171 & 2.2172 & 2.1372 & 2.2761 & 2.7332 \\
  \includegraphics[width=0.5in]{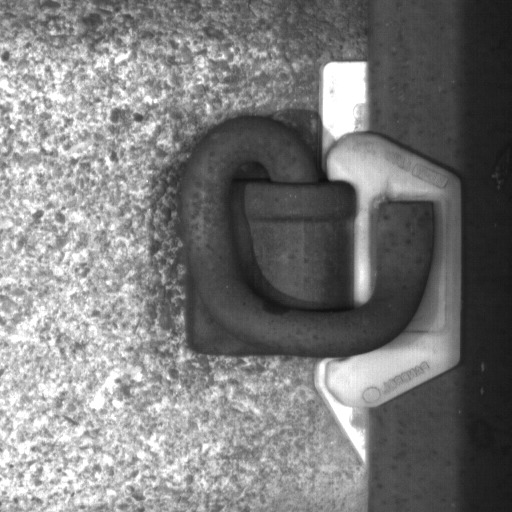} &
  \includegraphics[width=0.5in]{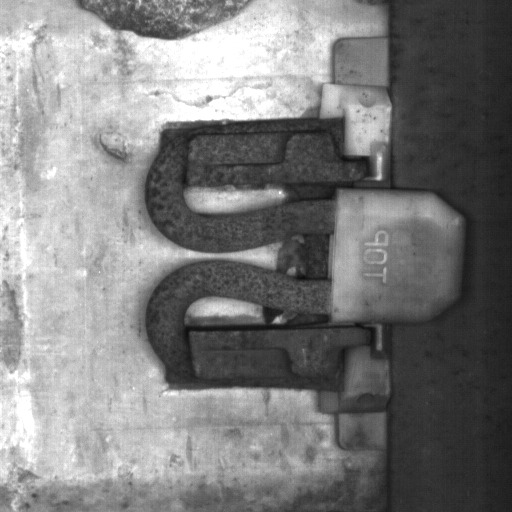} &
  \includegraphics[width=0.5in]{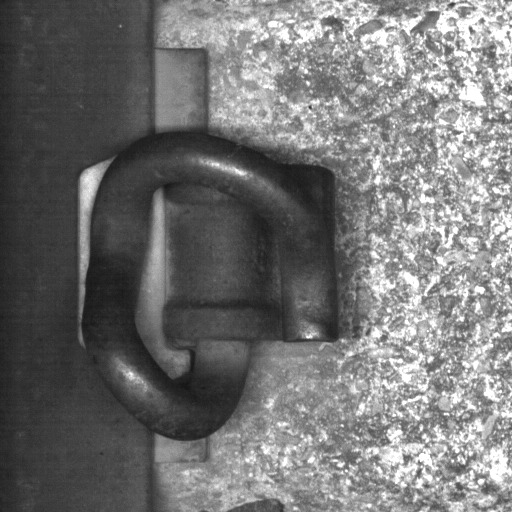} &
  \includegraphics[width=0.5in]{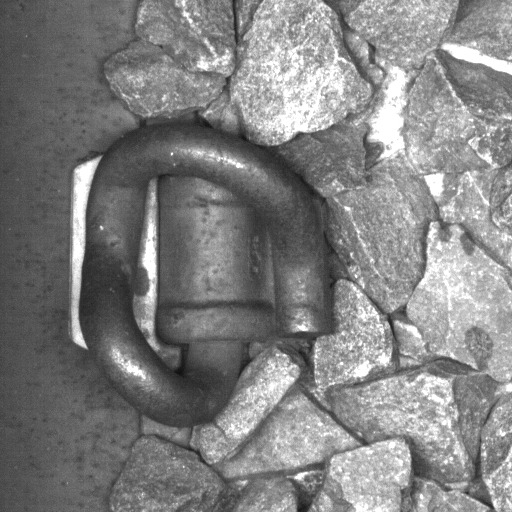} &
  \includegraphics[width=0.5in]{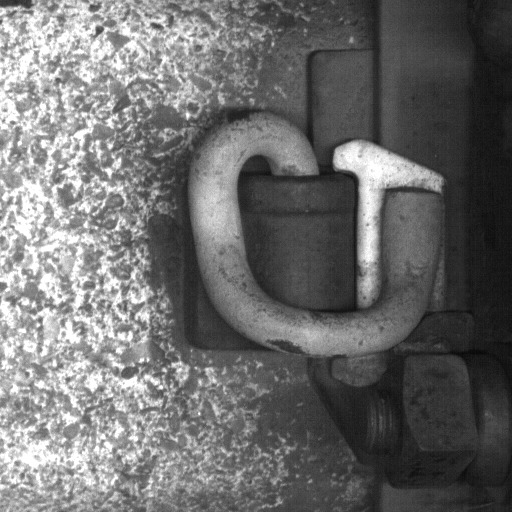} \\
  && (a) && \vspace{2mm} \\ 
  -1.5259 & -0.8281 & -0.7909 & -0.7995 & -0.5839 \\
  \includegraphics[width=0.5in]{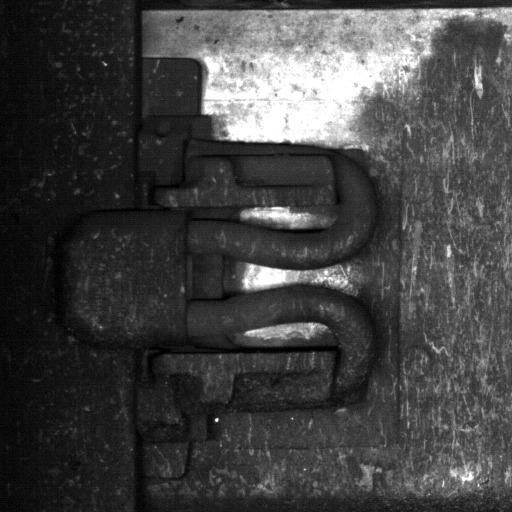} &
  \includegraphics[width=0.5in]{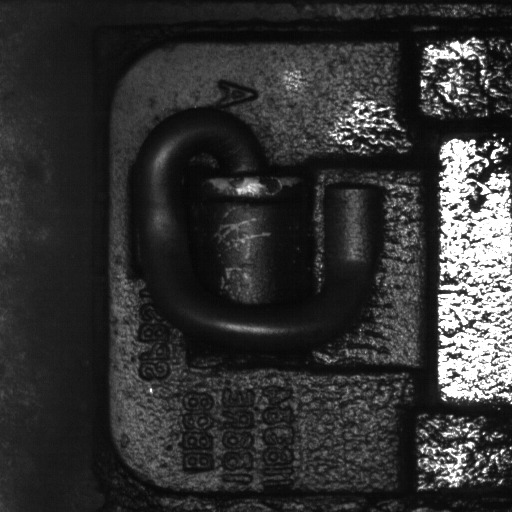} &
  \includegraphics[width=0.5in]{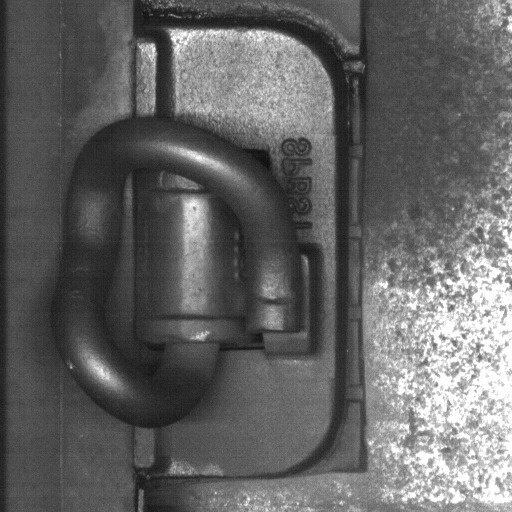} &
  \includegraphics[width=0.5in]{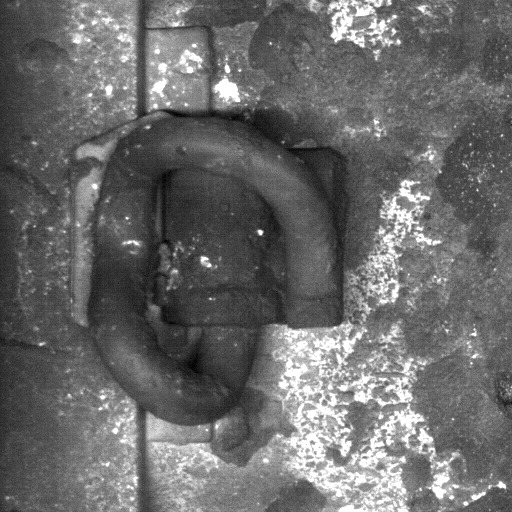} &
  \includegraphics[width=0.5in]{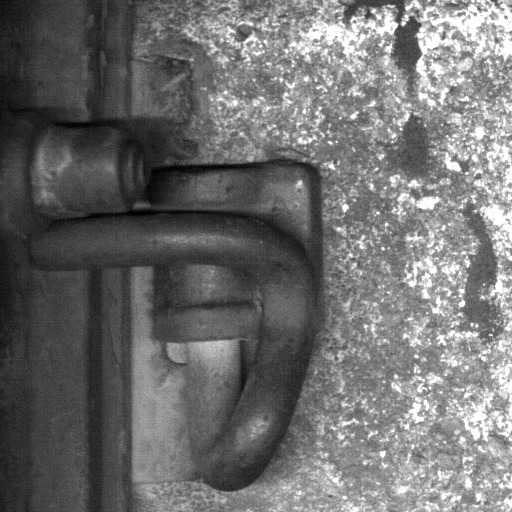} \\
  && (b) && \vspace{2mm} \\
  -0.2813 & -0.8813 & -0.8373 & -0.5157 & 1.4479 \\
  \includegraphics[width=0.5in]{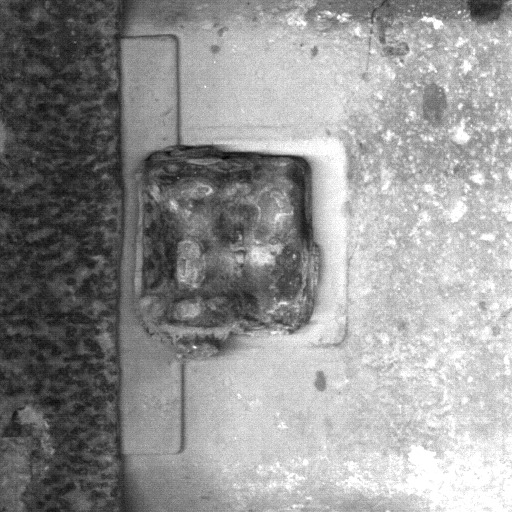} &
  \includegraphics[width=0.5in]{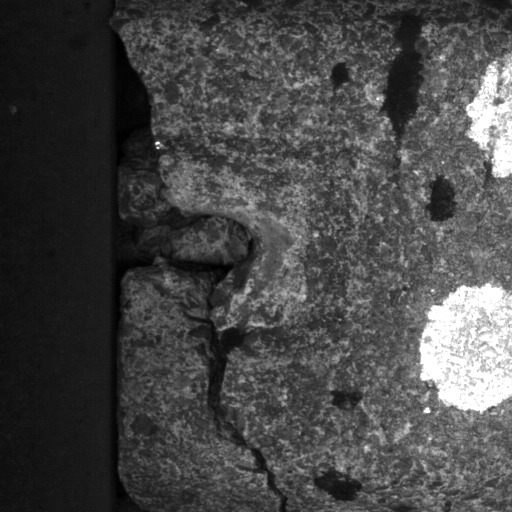} &
  \includegraphics[width=0.5in]{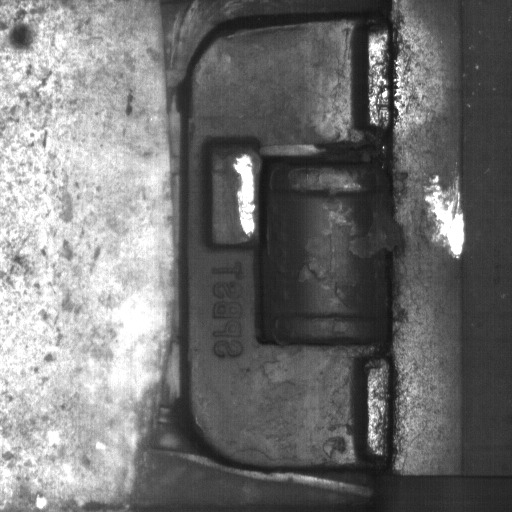} &
  \includegraphics[width=0.5in]{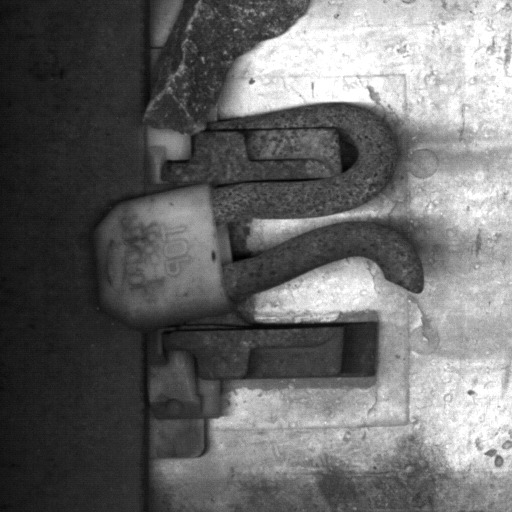} &
  \includegraphics[width=0.5in]{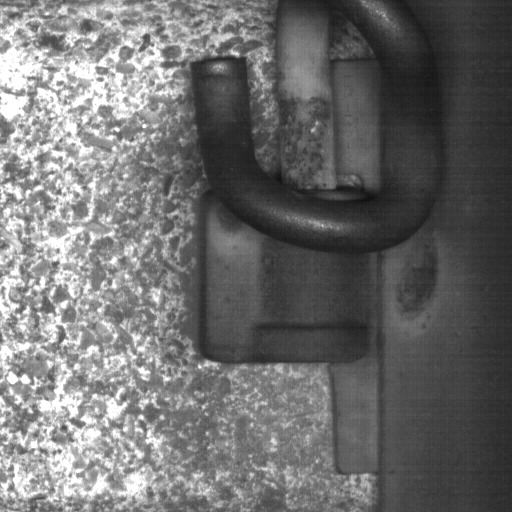} \\
  && (c) && \vspace{2mm} \\
  -2.0874 & -2.1373 & -2.3936 & -2.8944 & -2.5422 \\
  \includegraphics[width=0.5in]{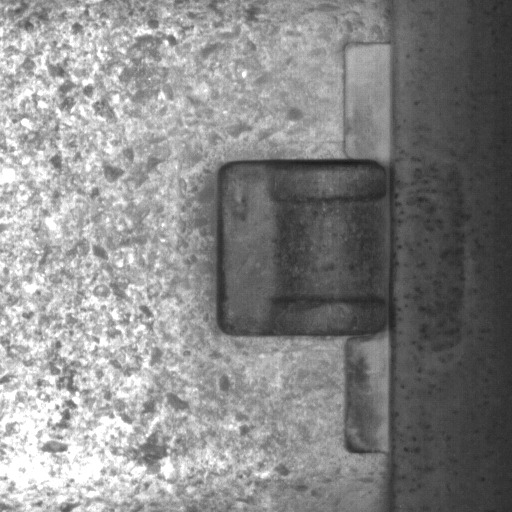} &
  \includegraphics[width=0.5in]{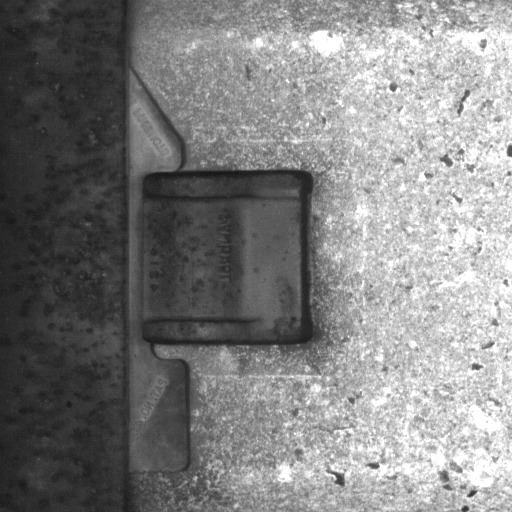} &
  \includegraphics[width=0.5in]{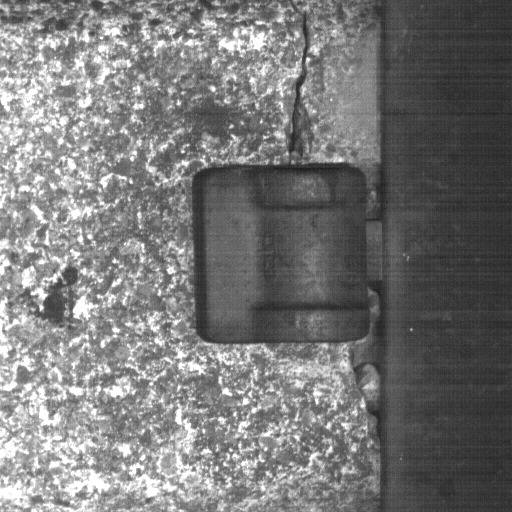} &
  \includegraphics[width=0.5in]{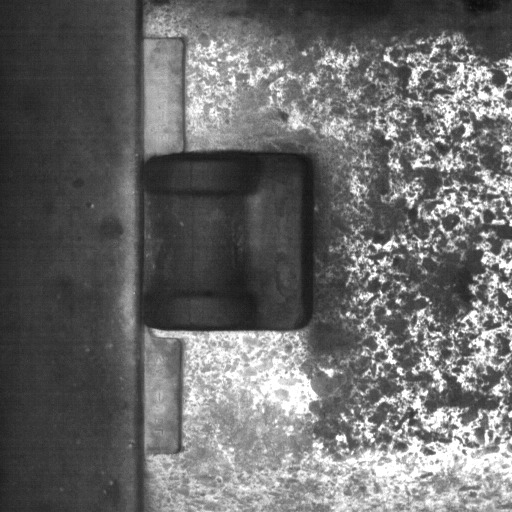} &
  \includegraphics[width=0.5in]{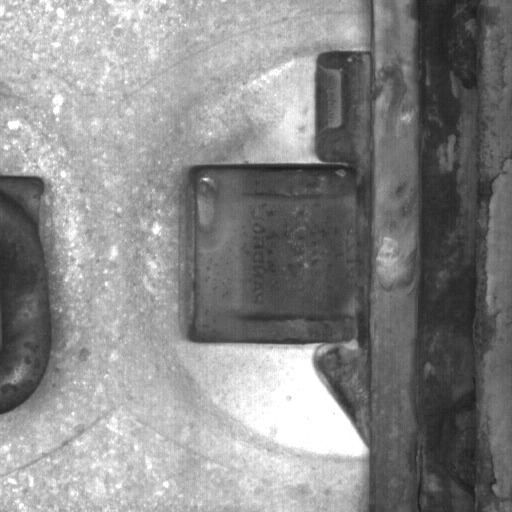} \\
  && (d) &&
\end{tabular}
\end{center}
\caption
{ \label{fig:example_scores}
Examples of fastener scores (a) Good fasteners with high scores (b) Good fasteners with low scores (c) Defective fasteners with high scores (d) Defective fasteners with low scores}
\end{figure}

Anomaly detection is a hypotheses testing problem in which the null hypothesis is that an image is normal and the alternative hypothesis is that it is anomalous. Due to the complexity of the scene and image formation process, both hypothesis are composite, with nuisance parameters arising from changes in illumination, occlusion, background clutter, and many other uncontrollable factors. Rather than trying to model each of these variables individually, in this paper we adapt the detection scores with the objective of reducing the variability in type I error rate. The is known as constant false alarm rate (CFAR) detection. We adopt the Bayesian view that such parameters are random variables with one realization per image. The images have a natural order based on the time they were captured at, so the sequence of these random parameters forms a random process. A key observation is that this random process has strong long-term dependencies. The effect of such slowly varying nuisance parameters is that false alarms are concentrated in small segments of the image sequence.

Figure \ref{fig:definitions} shows the definitions of several track components. In this paper, we focus on fastener inspection. Figure \ref{fig:example_scores} shows examples of good and defective fasteners and their detection scores generated by the multi-task learning method \cite{GP15multitask} of Gibert \etal. Although most fasteners have high scores and most defective ones have low scores, when good fasteners have low scores, there is an underlying phenomenon that causes scores of nearby images to also be low.

The rest of the paper is organized as follows. In Section~ \ref{sec:background} we review related works. The algorithm is described in Section~\ref{sec:approach}. Experimental results are described in Section~\ref{sec:results}. Section~\ref{sec:conclusion} concludes the paper with a brief summary and discussion.


\section{Background}\label{sec:background}

\subsection{Robust Anomaly Detection}

The presence of outliers is a challenge that many computer vision systems have to deal with. The RANdom SAmple Consensus (RANSAC) algorithm \cite{FB81} has been used in many applications for removing outliers when fitting a model to data. This method is specially useful when most of the samples follow a linear model plus additive i.i.d. Gaussian noise, but a few samples with gross errors do not follow this model. However, in many applications, it not clear which samples should be treated as inliers and which of them are outliers. For instance, in big data applications, the data just appears to have a distribution with long tails that decay at slower rate than the corresponding Gaussian distribution that best fits the data in the least squares sense. Indeed, what appears to be an outlier in feature space may just be a normal sample that has been subject to some kind of degradation for which the feature extractor was not designed for. These degradation modes may include impulse noise, partial occlusion, and in some cases, changes in appearance due to blur, shadows, or pose. In anomaly detection problems, the samples of interest are those in the tail of such data distribution. Therefore, any method that discards outliers have the potential of discarding anomalies, so in order to successfully find anomalies in such images it is necessary to use other methods.

The field of robust statistics \cite{Robust09,MMY06} provides the tools for estimation of unknown quantities when the underlying probability distribution is non-Gaussian and it is not known exactly. In practice, the data can be modeled as the mixture of a Gaussian distribution and a heavy-tailed distribution (the contaminated Gaussian model). In this case, it is desirable to design an estimator whose performance is minimax over a family of distributions that includes the Gaussian as a special case. There are basically three types of robust estimates: M-estimates\cite{H64} (Maximum likelihood type), L-estimates (Linear combination of order statistics), and R-estimates (Estimates derived from rank tests).

In supervised learning problems, there is a distinction on how to handle outliers at training time vs. testing time. Supervision at training time usually mitigates the problem of outliers as it is possible to manually select the inliers. The use of the $\ell_{1}$ minimization promotes a sparse representation of the data. The solution of the $\ell_{1}$ minimization is the Maximum Likelihood Estimate of the location parameter when the data follows a Laplacian distribution, and a straightforward way of robustifying a regression procedure is by replacing the $\ell_{2}$ norm in the cost function by the $\ell_{1}$ norm. A related L-estimator that results from such $\ell_{1}$ optimization is the Least Median of Squares (LMS), which was introduced in the computer vision field by Kim \etal \cite{RobustKKMMR89}. The drawback of the LMS is that the median estimator's efficiency is only $\frac{2}{\pi} = 0.637$ when the true distribution is Gaussian. The M-estimator based on the Huber loss function\cite{H64}
\begin{equation}
\rho(t) = \begin{dcases}
\frac{1}{2} t^2 & \text{for } |t| < k \\
k|t| - \frac{1}{2} k^2 & \text{for } |t| \geq k
\end{dcases}
\end{equation}
is more flexible because it has the sample mean ($k=\infty$) and sample median ($k=0$) as special cases and it can be tuned to handle different degrees of contamination in the contaminated Gaussian model. However, since this estimator depends on a scale parameter $k$ (unlike L-estimators, which are scale-invariant), it is necessary to first estimate this parameter using a robust scale estimator.

\subsection{Extreme Value Theory for Adaptive Anomaly Detection }

Due to illumination and viewpoint changes, clutter distribution, and other image degradation, the distribution of features extracted from images at test time, does not match what was observed during training. Moreover, such distribution may not be stationary, but slowly changes over time, so a fixed threshold would result in large variability in the false alarm rate. Broadwater and Chellappa\cite{BC10} proposed a technique to find adaptive thresholds for Constant False Alarm Rate (CFAR) detectors based on Extreme Value Theory (EVT) \cite{EG58} that can be used even when limited training data is available. EVT is applicable to problems where the probability of a rare event must be estimated even if such a rare event has never occurred. Scheirer \etal \cite{scheirer2010robust,scheirer2011meta} also used EVT for score normalization and showed its applicability to sensor fusion problems.

For completeness, we recall the EVT basic results below. Let $X_1,\dots,X_n$ be i.i.d. samples from an unknown distribution $F$ and $M_n = \max(X_1,\dots,X_n)$, the maximum of $n$ i.i.d. variables.  The fundamental EVT theorem, the Fisher-Tippett-Gnedenko theorem\cite{EG58}, states that if there exist a sequence of pairs of real numbers $(a_n,b_n)$ such that $a_n>0$ for all $n$ and a distribution function $\Lambda(x)$ such that
\begin{equation}
\lim_{n\rightarrow\infty} P \left( \frac{M_n - b_n}{a_n} \leq x \right) = \Lambda(x),
\end{equation}
for all $x$ at which $\Lambda(x)$ is continuous, then the limit distribution $\Lambda(x)$ belongs to either the Gumbel, the Fr\'{e}chet or the Weibull family. These three families can be grouped into the Generalized Extreme Value Distribution (GEVD)
\begin{equation}
\Lambda(x;\mu,\sigma,\xi) = \exp\left\{-\left[1+\xi\left(\frac{x-\mu}{\sigma}\right)\right]^{-1/\xi}\right\},
\end{equation}
where $\mu\in\mathbb R$ is the location parameter, $\sigma>0$ the scale parameter and $\xi\in\mathbb R$ the shape parameter. The Gumbel distribution is a special case of the GEVD when $\xi=0$, the Fr\'{e}chet when $\xi>0$, and the Weibull when $\xi<0$. When the limiting distribution exists, we say that $F(x)$ lies in the ``domain of attraction'' of $\Lambda(x)$.

In many practical applications, we are interested in the tail distribution of the distribution $F$. Given an upper threshold $u$, we select the $N_n$ samples that exceed such threshold and define the excesses $Y_1,\dots,Y_{N_n}$ as $Y_i = X_j - n$, where $i$ is the excess index and $j$ is the index of the original sample. The probability of exceeding the threshold is $\lambda = 1 - F(u)$. For sufficiently large $u$, the upper tail distribution function $F_u(y)$  (the conditional distribution function of the excesses),
\begin{equation}
F_u(y) = \frac{F(u+y) - F(u)}{1 - F(u)}
\end{equation}
can be approximated by a Generalized Pareto Distribution
\begin{equation}
G(y; \sigma,\xi) = 1 - \left( 1 + \frac{\xi y}{\sigma} \right)^{-1/\xi}_+, \quad y > 0.
\end{equation}
where $\sigma>0$, $\xi \in \mathbb{R}$, and $x_+ = \max(x,0)$. This approximation is justified by the Pickands theorem\cite{JP75}, which states that
\begin{equation}
\inf_\xi \lim_{u \uparrow \omega_F} \inf_{\sigma} \sup_{y>0} |F_u(y) - G(y; \sigma, \xi)| = 0
\end{equation} 
if and only if $F$ is in the domain of attraction of the GEVD. Note that the exponential distribution is a special case of the GPD for $\xi = 0$, i.e. $G(y;\sigma,0) = 1 - e^{-y/\sigma}$.

These results can be extended to the multivariate case, for example to model the tail distribution of the maximum of a cluster of observations. Under stationarity of observations, this can be achieved by incorporating both the tail of the marginal distribution and the so-called extremal index. Let $\{X_n : n \geq 1\}$ be a (strictly) stationary sequence of r.v.'s with marginal distribution $F$. Then, for sufficiently large $n$
\begin{equation}
P\{M_n \leq u_n \} \approx F^{n\theta}(u_n),
\end{equation}
where $u_n$ is any high threshold such that $n(1-F(u_n))$ converges to a positive number as $n\rightarrow \infty$ and $\theta$ is a fixed number in $[0,1]$. $\theta$ is the extremal index that measures the strength of dependence of $\{X_n\}$. If $\{X_n\}$ are independent, then $\theta=1$. On the other hand, if $\{X_n\}$ are highly dependent, then $\theta\approx 0$. A method for estimating the extremal index for a real-valued Markov chain was proposed by Yun \cite{SY98}. 


\section{Proposed Approach}\label{sec:approach}

In this section we describe our approach for normalizing the scores of an anomaly detector deployed in an application in which the distribution of the normal samples gradually changes over time. This may be caused by changes in illumination, change in view-point, addition or removal of clutter, or other uncontrollable factors. The approach is similar to the method proposed by Broadwater and Chellappa\cite{BC10} in which an adaptive threshold is estimated from the GPD fit to the upper tail of the distribution after removing the outliers or targets using a Kolmogorov-Smirnov statistical test. The difference is that our method is Bayesian and we work with sequential data and estimate the adaptive threshold for each sample.

\subsection{Bayesian Model}

We want to adapt the scores of an anomaly detector applied to a sequence of images so that, when we apply a given threshold, we get an approximate CFAR. The images have been collected from a moving vehicle, so the environmental conditions and clutter distribution are not stationary, but slowly change over time. In EVT-based threshold estimation, it is necessary to estimate the parameters $\sigma$ and $\xi$ of the GPD from the upper- or lower-tail of the empirical distribution. For the rest of this paper we will refer to the upper tail of the distribution of the random variable $X$, but the same applies to the lower tail since the lower tail of $X$ is the upper tail of $Z = -X$. The threshold $u$ needs to be set high enough so that the tail of $F(x)$ converges in distribution to the GPD. However, since we are dealing with a non-stationary random process, we need to work on a small window centered at the sample of interest. This window needs to be long enough so that we can fit the parameters of the GPD to its tail (for example the largest 5\% of the samples), but short enough that the distribution has not changed much. In applications in which the dynamics of the process change quickly, our options are rather limited. If we fit a GPD to the extreme samples of a short window, the estimated threshold has so much variance that the resulting performance is worse than using a fixed threshold. On the other hand, if the window is too long, the threshold does not adapt at all. For example, if we use a window of 100 samples and select the upper threshold to the 95th percentile, we would only have 5 samples to estimate the 2 parameters of the GPD, resulting in severe overfitting.

To overcome this limitation, we will make one simplification by fixing $\xi = 0$, so we only need to estimate one parameter instead of two. Under $\xi = 0$, the GPD reduces to the exponential distribution
\begin{align}
G(y;\sigma,\xi=0) = 1 - e^{-y/\sigma}.
\end{align}
For convenience, we apply the parameterization $\lambda = 1/\sigma$ and write the Exponential in its canonical form
\begin{align}
G(y;\lambda) &= 1 - e^{-\lambda y}\label{eq:exp} \\
g(y;\lambda) &= \lambda e^{-\lambda y}.
\end{align}
As opposed to the general case of the GPD, the Exponential distribution is a member of the exponential family, so it has a non-trivial sufficient statistic from which we can easily compute the maximum likelihood estimate (MLE) of its parameter. Its conjugate prior is the Gamma distribution,
\begin{align}
\pi(\lambda;\alpha, \beta) = \frac{\beta^\alpha}{\Gamma(\alpha)} \lambda^{\alpha-1} e^{-\beta \lambda}\text{,}
\end{align}
the non-informative (improper) prior is given by $\alpha=1$, $\beta = 0$, and the parameters of the Gamma posterior under a $\text{Gamma} (\lambda;\alpha_0,\beta_0)$ prior can be computed as
\begin{align}
\alpha_1 &= \alpha_0 + n\\
\beta_1 & = \beta_0 + \sum_{i=1}^n y_i.
\end{align} Moreover, the maximum a posteriori (MAP) estimate has the closed from $\widehat{\lambda}=\frac{\beta}{\alpha-1}$. This simplified model allows us to derive a very fast adaptation algorithm that we describe in the following section. This approximation works well in practice, especially when the scores are trained with a sparsity promoting loss function such as the hinge loss.

\subsection{Training}
Our training set $\mathcal{T}$ contains a number of sequences of scores $\vec{x}$ with their corresponding sequences of labels $\vec{y}$. During training, we compute the sufficient statistics $n$ and $s$ for all the samples that are not labeled as anomalies (the sufficient statistic is all we need to characterize the Gamma prior distribution). We then re-scale them to limit the effect of this prior. Effectively, we use $w_0$ pseudo-samples instead $n$ (the number of samples in the training set). This is necessary because $n$ is usually a very large number, and computing $\alpha_0$ and $\beta_0$ with it would result in a very strong prior that would introduce too much bias in the MAP estimate.

The steps of the training procedure are described in Algorithm~\ref{alg:trainevt}. The parameter $p_u$ is the probability of the tail, and $w_0$ is the weight in sample counts that we assign to the training set. In our experiments we used $p_u = 0.05$ and $w_0 = 400$.

\begin{algorithm}[tp!]
\caption{EVT training algorithm.}
\label{alg:trainevt}
\begin{algorithmic}[1]
\Procedure {Train}{$\mathcal{T}$, $p_u$, $w_0$}
  \State $n \gets 0$, $s \gets 0$	\Comment{Initialize sufficient statistics}
  \ForAll {$(\vec{x},\vec{y}) \in \mathcal{T}$}	\Comment{Training set $\mathcal{T}$} contains $\vec{x}$ scores, $\vec{y}$ labels
    \State $\vec{g} \gets \{x_i \mid y_i = 0\}$	\Comment{Select negative samples}
    \State $u \gets u \mid \#\{g_i > u\} = \#\vec{g} \; p_u$	\Comment{Find upper threshold}
    \State $\vec{t} \gets \{g_i \mid g_i > u \}$ - u		\Comment{Extract upper tail}
    \State $n \gets n + \#\vec{t}$		\Comment{Update counts}
    \State $s \gets s+ \sum\vec{t}$	\Comment{Update sum}
  \EndFor
  \State $\alpha_0 \gets 1 + w_0$
  \State $\beta_0 \gets \frac{w_0  \; s} {n}$
  \State \textbf{return} $\alpha_0$, $\beta_0$	\Comment{Parameters of the Gamma prior}
\EndProcedure
\end{algorithmic}
\end{algorithm}

\begin{algorithm}[tp!]
\caption{EVT adaptive thresholding algorithm}
\label{alg:testevt}
\begin{algorithmic}[1]
\Procedure {AdaptScores}{$\vec{x}$, $\alpha_0$, $\beta_0$, $p_u$, $p_f$, $w_1$, $L$, $n_a$}
  \State $\widehat{\lambda}_0 \gets \frac{\beta_0}{\alpha_0 - 1}$	\Comment{MLE in training set}
  \State $\vec{y} \gets \textbf{sort\_desc}(\vec{x})$	\Comment{Sort scores in descending order}
  \State $k \gets \#\vec{y} \; p_u$
  \For{$i \gets 1, n_a$}	\Comment{Training set $\mathcal{T}$} contains $\vec{x}$ scores, $\vec{y}$ labels
    \State $u \gets y_{i+k}$	\Comment{Find upper threshold}
    \State $\vec{t} \gets \{y_i, \dots, y_{i+k}\} - u$		\Comment{Extract upper tail}
    \State $D_{n,i} = \sup_{x \in \vec{t}} \left|\widehat{G}_n(x) - G(x;\lambda)\right|$	\Comment{Compute KS statistic}
  \EndFor
  \State $\hat{i} \gets \min_i \{ D_{n,i} \}$	\Comment{Estimate number of outliers}
  \State $u' \gets y_{\hat{i}}$	\Comment{Set outlier rejection threshold}
  \State $\vec{t} \gets \{y_{\hat{i}} , \dots, y_{\hat{i}+k}\} - u$	\Comment{Extract upper tail}
  \State $\alpha_1 \gets \alpha_0 + w_1$
  \State $\beta_1 \gets \beta_0 + \frac{w_1  \; \sum \vec{t}} {\#\vec{t}}$
  \For{$i \gets 1, n$}
    \State $\vec{w} \gets \vec{x}_{i-(L-1)/2:i+(L-1)/2}$	\Comment{Window centered at sample $x_i$}
    \State $u \gets u \mid \#\{w_i > u\} = \#\vec{w} \; p_u$	\Comment{Find upper threshold}
    \State $\vec{t} \gets \{w_i \mid w_i > u \}$ - u		\Comment{Extract upper tail}
    \State $\alpha \gets \alpha_1 + \#\vec{t}$		\Comment{Posterior}
    \State $\beta \gets \beta_1+ \sum\vec{t}$	\Comment{Posterior}
    \State $\widehat{\lambda} \gets \frac{\beta}{\alpha-1}$	\Comment{MAP estimate}
    \State $y_i \gets x_i + u - \widehat{\lambda} \; log(p_f/p_u)$	\Comment{Adapt score}
  \EndFor
  \State \textbf{return} $\vec{y}$		\Comment{Adapted scores}
\EndProcedure
\end{algorithmic}
\end{algorithm}



\subsection{Proposed Adaptive Thresholding Algorithm}
During testing, we first perform a series of Kolmogorov-Smirnov (KS) tests \cite{kendall99} to find and remove anomalies. The KS statistic
\begin{align}
D_n = \sup_x \left|\widehat{G}_n(x) - G(x; \widehat{\lambda})\right|
\end{align}
measures the dissimilarity between distributions $G(x;\widehat{\lambda})$ and $\widehat{G}_n(x)$. $G(x;\widehat{\lambda})$ is the GPD in \eqref{eq:exp} and
\begin{align}
\widehat{G}_n(x) = 1 - \frac{1}{n} \sum_{i=1}^{n} I(X_i \leq x) \label{eq:empirical}
\end{align}
where $I(x)$ is a standard indicator function, is the empirical tail CDF. The KS test requires estimating a threshold $K_\alpha$ for rejecting (with confidence $1-\alpha$) the hypothesis that the observed data does not fit G with the test $\sqrt{n} D_n > K_\alpha$. The result from Lilliefors\cite{lilliefors67} shows that the KS test is biased when the reference distribution G is not precisely known (in this case, $\widehat{\lambda}$ is estimated from the training data). However, as noted in \cite{BC10}, it is not necessary to identify the exact value of $\alpha$ for the purpose of removing anomalies and outliers. Instead, we first compute $D_n$ with all the samples in the tail. We call this $D_{n,1}$. We then remove the largest sample and we compute $D_{n,2}$ using the remaining samples. We keep iterating until we get $D_{n,n_a}$. Finally, we select the value of $i$ that minimizes $D_{n,i}$.

After removing the anomalies, we use the prior estimated during training to compute the posterior for the whole sequence. This posterior is used as the prior for estimating the tail distribution on each shift of a window centered on each of the samples. The details of the adaptation procedure are described in Algorithm~\ref{alg:testevt}. The input to the adaptation procedure is a sequence of scores $\vec{x}$, the parameters of the prior Gamma distribution $\alpha_0$ and $\beta_0$, the size of the upper tail $p_u$, the target false alarm rate $p_f$, the weight $w_1$ that we assign to the the prior contribution of the whole sequence, the window length $L$, and the maximum number of anomalies $n_a$ in the sequence.  The output sequence $\vec{y}$ has been adapted so that when it is thresholded at 0, the false alarm rate is $p_f$. In our experiments, we have used $p_u = 0.05$, $p_f=0.001$, $w_1 = 100$, $L=101$, and $n_a=12$.

\section{Experimental Results}\label{sec:results}

To validate the effectiveness of the proposed approach, we have used the 340 sequences of fastener detections corresponding to each of the 4 cameras in each of the 85 miles of the Amtrak NEC concrete tie dataset introduced in \cite{GP15fast}. This dataset contains a total of 203,287 ties and each tie is divided in 4 regions (left field, left gage, right gage, and right field), so the total number of images is 813,148. The detection problem consists in determining whether an image contains a fastener attached to one of the rails. The dataset contains bounding boxes for all the images that are known to contain a defect. The total number of defects is 1,087 (0.13\% of all the fasteners). The defective fastener class contains two subclasses: broken fastener and missing fastener.

We have used the scores generated by the multi-task learning (MTL) detector described in \cite{GP15multitask}. This detector uses deep learning with multiple tasks that are trained in parallel. The reason for using multiple tasks is to prevent overfitting. By sharing a common low-level representation between the fastener inspection task and a separate material classification task, there is a data amplification effect that results in better generalization for both classifiers. We also compare the performance with the baseline single-task learning (STL) method in \cite{GP15fast}. The raw data was provided by Amtrak, and the authors of \cite{GP15multitask,GP15fast} provided the output of their detectors as well as the codes to evaluate the performance. This detector produces a scalar-valued score for each image by spatially pooling all the detections in the image. Scores are high when the image contains a good fastener, and low when the fastener is either missing or broken. Figure \ref{fig:example_scores} shows several detection examples of the MTL detector.

%

\begin{figure}[htp!]
\begin{center}
  \includegraphics[width=3.25in]{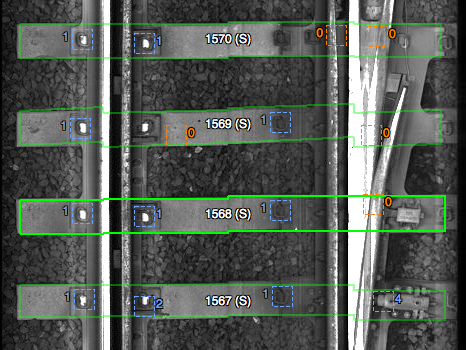}
\end{center}
\caption
{ \label{fig:def_switch}
Example of section marked as switch.}
\end{figure}

\begin{figure}[htp!]
\begin{center}
  \includegraphics[width=3.25in]{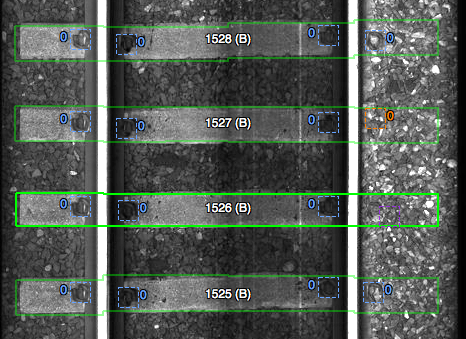}
\end{center}
\caption
{ \label{fig:def_ballast}
Example of section marked as ballast.}
\end{figure}

\begin{figure}[htp!]
\begin{center}
  \includegraphics[width=3.25in]{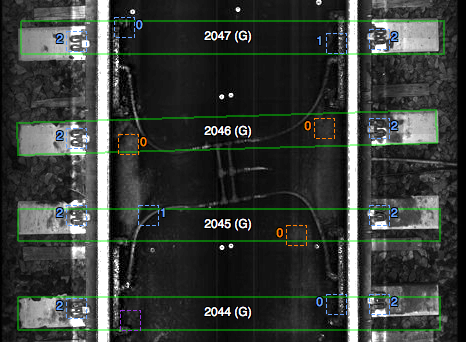}
\end{center}
\caption
{ \label{fig:def_lubricator}
Example of section marked as lubricator.}
\end{figure}

To facilitate the evaluation of fastener detection performance under difficult scenarios, whenever the fastener is not directly attached to the rail or tie, or when for some reason a fastener is not visible at all, those ties are marked as uninspectable with a special label. Depending on the value of such label, the dataset is divided into 3 subsets: 
\begin{itemize}
\item \emph{Clear ties:} 200,763 ties (1,037 ties with at least one defect).
\item \emph{Clear ties plus switches:} 201,856 ties (1,045 ties with at least one defect). See Figure~\ref{fig:def_switch} for an example of a switch section.
\item \emph{All ties:} 203,287 ties (1,052 ties with at least one defect). This includes switches, and ties for which some fasteners are not visible because they are covered by ballast or a lubricator. See Figures \ref{fig:def_ballast} and \ref{fig:def_lubricator} for examples of high ballast and lubricator sections. 
\end{itemize}

\begin{figure*}[htp!]
 \centering
\includegraphics[width=11cm]{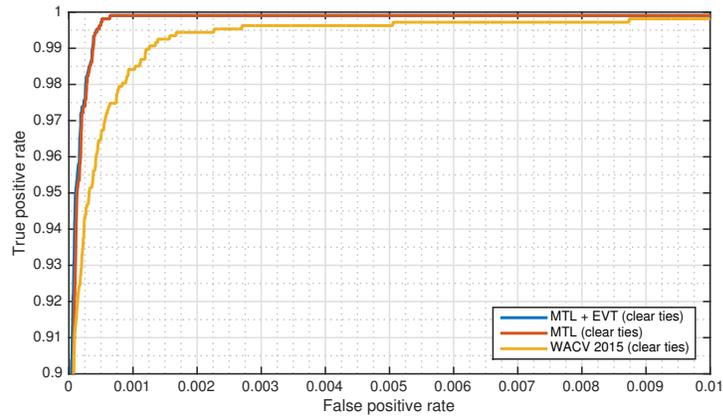}\\
(a)\\
\includegraphics[width=11cm]{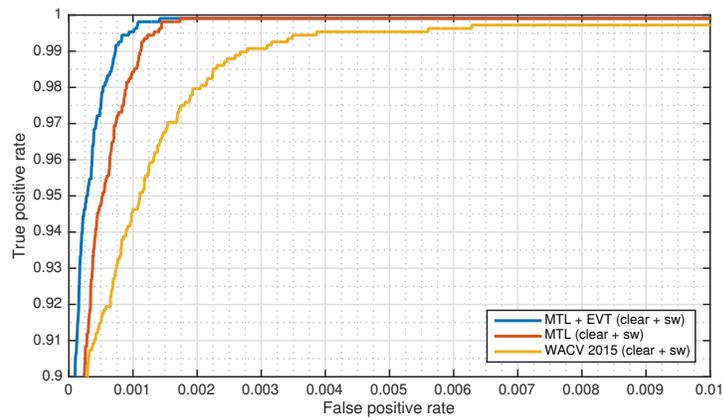}\\
(b)\\
\includegraphics[width=11cm]{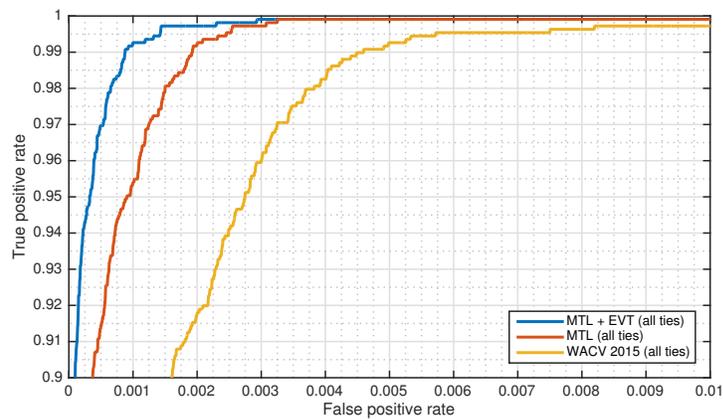}\\
(c) \vspace{4mm}\\
\caption{ROC curves comparing defective fastener detection performance on the 85-mile testing set using normalized vs. unnormalized scores (a) on the clear ties subset (b) on the clear with with switches subset (c) on all ties. Detections are per image (each tie has 4 images).}
\label{fig:roc_mtl}\vspace{2mm}
\end{figure*}

For training, we use all the available data after setting aside the sequence being tested. Table~\ref{table:condition_results} and Figure~\ref{fig:roc_mtl} show the detection results on the normalized scores. The overall improvement is significant. The detection rate on the whole dataset at $PFA = 0.1\%$ increases from $95.40\%$ to $99.26\%$. This is a $6\times$ reduction in the missed rate. Moreover, the performance on the clear tie subset does not degrade at all. The running time of our EVT adaptation algorithm implemented in MATLAB\footnote{The code and data used in this section is available at\\ \url{https://github.com/xavigibert/EvtTrack}} for adapting all 813,148 scores is only of 17 seconds on a Mid-2012 MacBook Pro with a 2.5 GHz Intel Core i5 processor, so this dramatic improvement comes at negligible computational cost (running the detector process takes several hours).

\section{Conclusions}\label{sec:conclusion}

In this paper, we presented a new algorithm that normalizes scores from a sequential anomaly detector with the objective of harmonizing its false alarm rate. Extreme value theory provides a solid foundation from which adaptive thresholding algorithms can be derived. When working with sequences of images, we need to take advantage of the statistical dependencies of nuisance parameters of nearby images. If we discard such dependencies and treat each image in the sequence independently, the performance suffers.

The CFAR detection approach proposed in this paper has applicability beyond railway track inspection from a moving vehicle. It could be used, for example, in surveillance video to remove bursts of false alarms caused by sun glare, insects, rain or fog. Its computational cost is negligible compared to that of the underlying detector, so this approach can be easily retrofitted to existing detectors already in operation.

\begin{table*}[htp!]
\begin{center}
\begin{tabular}{c | c | c | c | c }
  \hline
  Condition & PFA & MTL + EVT & MTL\cite{GP15multitask} & STL\cite{GP15fast} \\
  \hline
  \hline
  \multirow{2}{*}{Fastener (only clear ties)} & 0.1\% & \textbf{99.91\%} & \textbf{99.91}\% & 98.41\% \\
  & 0.02\% & \textbf{97.20}\% & 96.74\% & 93.19\% \\
  \hline
  \multirow{2}{*}{Fastener (clear + switch)} & 0.1\% & \textbf{99.54}\% & 98.43\% & 94.54\% \\
  & 0.02\% & \textbf{93.80}\% & 89.35\% & 88.70\% \\
  \hline
  \multirow{2}{*}{Fastener (all ties)} & 0.1\% & \textbf{99.26}\% & 95.40\% & 87.38\% \\
  & 0.02\% & \textbf{93.47}\% & 87.76\% & -- \\
  \hline
\end{tabular}
\end{center}
\caption{\label{table:condition_results}Fastener detection results before and after score normalization.}
\end{table*}

\section*{Acknowledgements}
The authors thank Amtrak, ENSCO, Inc. and the Federal Railroad Administration for providing the data used in this paper.

{\small
\bibliographystyle{ieee}
\bibliography{evtbib}
}

\end{document}